\documentclass[11pt]{article}

\usepackage[preprint]{acl}

\usepackage{times}
\usepackage{latexsym}
\usepackage{amsmath}
\usepackage{amssymb}

\usepackage[T1]{fontenc}

\usepackage[utf8]{inputenc}

\usepackage{microtype}

\usepackage{inconsolata}

\usepackage{graphicx}
\usepackage{makecell}
\usepackage{multirow}
\usepackage{enumitem}

\title{Building Data-Driven Occupation Taxonomies: A Bottom-Up Multi-Stage Approach via Semantic Clustering and Multi-Agent Collaboration}

\author{Nan Li \and Bo Kang \and Tijl De Bie \\
    IDLab, Department of Electronics and Information Systems \\
    Ghent University, Belgium}

\begin{document}

\maketitle

\begin{abstract}
Creating robust occupation taxonomies, vital for applications ranging from job recommendation to labor market intelligence, is challenging.
Manual curation is slow, while existing automated methods are either not adaptive to dynamic regional markets (top-down) or struggle to build coherent hierarchies from noisy data (bottom-up). We introduce CLIMB (CLusterIng-based Multi-agent taxonomy Builder), a framework that fully automates the creation of high-quality, data-driven taxonomies from raw job postings. CLIMB uses global semantic clustering to distill core occupations, then employs a reflection-based multi-agent system to iteratively build a coherent hierarchy. On three diverse, real-world datasets, we show that CLIMB produces taxonomies that are more coherent and scalable than existing methods and successfully capture unique regional characteristics. We release our code and datasets at \url{https://anonymous.4open.science/r/CLIMB}.
\end{abstract}

\section{Introduction}

A taxonomy is a hierarchical structure that organizes information. In the labor market, a robust taxonomy is critical for organizing job postings, guiding job seekers, and informing policy. 
It improves search and guidance for job seekers, provides governments a framework to analyze labor trends and inform policy, and underpins corporate workforce planning and skill gap analysis.
However, because labor markets are dynamic and regionally diverse, there is a strong need for taxonomies that are tailored to specific markets and easily updatable. The goal of this work is to fully automate the construction of such data-driven, hierarchical taxonomies directly from a raw corpus of job postings.

Existing methods struggle to meet this need. Manual curation is slow, expensive, and unscalable. As detailed in Section~\ref{sec:related-work}, automated approaches are either \emph{top-down}, expanding an existing structure, or \emph{bottom-up}, building a new one from data.

Many traditional and recent LLM-based methods follow a top-down approach. These systems are designed to expand upon a pre-existing \emph{seed}, typically a small, expert-curated taxonomy or a list of initial terms extracted and filtered using classic natural language processing techniques. While effective for enriching existing knowledge structures, this reliance on a seed makes them ill-suited for building a taxonomy from scratch in a new domain and and fails to fully remove the need for expert knowledge and potential for human bias.

In contrast, bottom-up approaches aim to build a hierarchy directly from a raw corpus but face fundamental challenges. \emph{First}, they struggle to distill 
a globally consistent set of core concepts (e.g., distinct occupations) from the data;
 feeding an entire corpus to an LLM is often infeasible, while incremental processing sacrifices the global perspective of the whole corpus needed to identify concepts consistently. \emph{Second}, even with a clean set of concepts, constructing a deep and logically coherent hierarchy is a complex reasoning task where single-LLM systems often fail, producing inconsistent groupings and flawed structures.

\begin{figure}[t]
    \centering
    \includegraphics[trim=0mm 0 1mm 0, clip, width=0.5\textwidth]{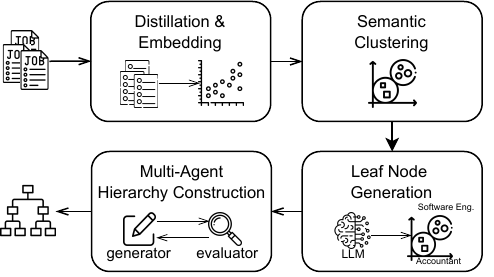}
    \caption{High-level overview of CLIMB.}
    \label{fig:pipeline}
\end{figure}

To address these challenges, we propose CLIMB (\textbf{CL}uster\textbf{I}ng-based \textbf{M}ulti-agent taxonomy \textbf{B}uilder), a multi-stage framework that solves both problems in sequence. To overcome the first challenge, CLIMB begins with the raw corpus and uses semantic clustering over the entire dataset to automatically distill a globally-informed set of leaf-node concepts. To solve the second, CLIMB employs a novel reflection-based multi-agent framework for the hierarchical construction. It proceeds level-by-level, using a ``Generator'' to propose parent concepts and an ``Evaluator'' to critique them for logical consistency. This iterative refinement is essential for building a complex yet coherent taxonomy. Our \emph{contributions} are: 
\begin{itemize}[leftmargin=*, noitemsep, topsep=0pt]
    \item We propose a novel multi-stage methodology, CLIMB, that fully automates bottom-up taxonomy generation. Its key innovations are (1) a global clustering stage to distill concepts directly from a raw corpus and (2) a multi-agent, reflection-based framework to solve the complex reasoning task of hierarchical construction. This makes the process objective by removing the need for expert seeds or manual curation.
    \item We demonstrate that CLIMB produces taxonomies that are not only coherent and scalable, but also highly adaptive. Unlike static, generic taxonomies, CLIMB's data-driven approach creates structures that are customized to specific regional labor markets at a specific point in time, capturing unique local roles and emerging trends.
    \item We release our code and the datasets used in this study to the public to encourage reproducible research and further innovation in this area (\url{https://anonymous.4open.science/r/CLIMB}).
\end{itemize}

\section{Related Work}\label{sec:related-work}

The task of taxonomy generation is broadly divided into two main methods: top-down and bottom-up.

\textbf{Top-Down Approaches.} A significant body of work, from both the pre-LLM era \cite{shen2018hiexpan,zhang2018taxogen,huang2020corel,shang2020nettaxo,lee2022taxocom,le2023automatic} and the recent LLM era \cite{zeng2024chain,gunn2024creating,marchenko2024taxorankconstruct,kargupta2025taxoadapt}, operates top-down by expanding a pre-existing seed taxonomy. While effective for enriching established knowledge structures, this reliance on a seed makes them less suitable for creating a taxonomy from scratch in a new domain.

\textbf{Bottom-Up Approaches.} In contrast, bottom-up approaches aim to construct a taxonomy directly from a corpus, aligning with CLIMB's objective of creating data-driven, adaptive taxonomies. 
Recent works have explored several strategies to this end. Some domain-specific methods require a pre-compiled list of terms \cite{sas2024automatic,moraes2024using}, precluding full automation from raw text. 
To the best of our knowledge, the state-of-the-art method that operates in a fully automated, bottom-up, and seed-free manner is TnT-LLM \cite{wan2024tnt}, which we select as our primary baseline. TnT processes data in small batches to iteratively build its taxonomy. 
While scalable, this local view (i.e., seeing only the data in the current batch) can result in redundant concepts or a fragmented hierarchy, especially with noisy data. 
Another strategy uses a human-in-the-loop to refine an LLM-generated draft, sacrificing full automation for quality control \cite{shah2023using}. In contrast, CLIMB's fully automated solution resolves the local-view limitation by performing global clustering on the corpus, ensuring its foundation of leaf nodes is comprehensive and consistent.

\textbf{Reflection for Complex Reasoning.} Once base concepts are identified, constructing a logically sound hierarchy is a complex reasoning task where a single LLM struggles to ensure overall logical coherence. In other fields require such reasoning, such as code generation \cite{madaan2023self,chen2023teaching} or mathematical problem-solving \cite{yao2023tree}, reflection-based multi-agent frameworks using a ``generator'' and ``evaluator'' for iterative refinement have proven highly effective \cite{anthropic2024building,kalyanpur2024llm,yuan2025reinforce}. To our knowledge, CLIMB is the \emph{first} to apply this powerful paradigm to coherent and robust taxonomy construction.

\section{Method}\label{sec:method}

Given a corpus of documents, our goal is to automatically construct a hierarchical taxonomy.
As aforementioned, building a taxonomy from a raw corpus has two challenges: (1) distilling a consistent set of base concepts from a large and noisy corpus, and (2) constructing a coherent hierarchy from them. CLIMB's multi-stage pipeline (Figure~\ref{fig:pipeline}) is designed to address these.
\emph{First}, the pipeline generates leaf nodes from raw job postings. This involves optional text distillation and embedding (Section~\ref{sec:distillation}), global semantic clustering to identify occupations (Section~\ref{sec:clustering}), and then transforming raw clusters into canonical leaf nodes via LLM-based abstraction and normalization (Section~\ref{sec:leaf-node-labeling}).
\emph{Second}, for hierarchical construction (Section~\ref{sec:hierarchical}), a reflection-based multi-agent framework iteratively builds a logically coherent taxonomy upwards from the leaves, level by level. Implementation details are provided in Appendix~\ref{app:implementation-details}.

\subsection{Job Posting Distillation \& Embedding}\label{sec:distillation}

Job postings often mix core occupational details with irrelevant text about company culture or application procedures. Applying a large LLM to summarize an entire corpus is prohibitively slow and expensive. 
We therefore introduce an optional, scalable distillation step: an LLM cost-effectively generates a high-quality training set for a lightweight classifier, which then extracts relevant text segments from all postings for subsequent embedding. The process involves the following key steps:

\noindent \emph{Data Preparation}: All job descriptions first undergo basic text cleaning (e.g., removing HTML tags and extra whitespace) and are then segmented into text chunks for relevance classification.

\noindent \emph{LLM Labeling}: To create training data cost-effectively, an LLM annotates a sample of chunks as relevant or irrelevant to the core occupation.

\noindent \emph{Classifier Training}: A lightweight binary classifier is trained on the LLM-annotated data, enabling scalable filtering of all job postings.

\noindent \emph{Distilled Description Generation}: The classifier extracts relevant chunks from each posting to create a ``distilled'' description. For robustness, if distillation is not needed or yields no relevant text, the full, preprocessed posting is used as a fallback.

\noindent \emph{Embedding}: The resulting descriptions are embedded using a pre-trained language model for the subsequent clustering stage.

\subsection{Semantic Clustering}\label{sec:clustering}
The next step is to group distilled job descriptions into fine-grained occupation clusters (leaf nodes of a taxonomy tree). Since simple embedding cosine similarity fails to capture the nuanced human judgment of what constitutes the ``same occupation,'' we learn a specialized similarity metric by training a classifier to mimic an HR expert's assessment. The process involves the following steps:

\noindent \emph{Contrastive Data Sampling}: To create a rich training set, we sample pairs using a coarse initial signal from general-purpose embedding cosine similarity. This is \emph{not} a labeling step, but a strategy to deliberately select a challenging mix for LLM annotation: likely easy-positives (high similarity), likely easy-negatives (low similarity), and likely ambiguous hard-positives and -negatives (moderate similarity).

\noindent \emph{LLM as HR Expert}: We employ a powerful, general-purpose LLM to act as a proxy for an HR expert, a common and proven effective strategy for scalable data annotation\cite{gilardi2023chatgpt,zheng2023judging}, to provide ground-truth ``same occupation'' labels for the sampled pairs. This supervision forces the classifier to learn from these ambiguous pairs and develop a nuanced similarity metric that surpasses the initial embeddings. The prompts are available in our public repository.

\noindent \emph{Classifier Training}: An XGBoost classifier is trained on these LLM-generated labels. The feature set for a pair of job embeddings, \(e_a\) and \(e_b\), is the concatenation of \((e_a, e_b, e_a - e_b, e_a \odot e_b)\), capturing rich interaction information.

\noindent \emph{Clustering with Learned Similarity}: The trained classifier is used to compute a similarity score for all job pairs. These scores then serve as the input for the Affinity Propagation algorithm (selected in early tests for its superior silhouette scores), to group postings into distinct occupation clusters.

\subsection{Leaf Node Generation}\label{sec:leaf-node-labeling}
The raw clusters from the previous step are groups of job postings, but their varied and often inconsistent titles are unsuitable for a formal taxonomy. This stage transforms these clusters into clean, canonical occupations with clear titles and descriptions, serving as the leaf nodes of the taxonomy.

\noindent \emph{LLM-based Abstraction}: We prompt an LLM with the cluster's (sampled if too many) job postings to generate a concise title and description for the occupation each cluster represents, crucial for abstracting a canonical concept from noisy raw text. 

\noindent \emph{Normalization and Deduplication}: LLM-generated titles can be ambiguous or redundant, so we refine them as follows. (1)
Ambiguous conjoined titles (e.g., ``Accountant and Bookkeeper'') are removed to ensure one concept per node.
(2) Standard text cleaning is applied.
(3) Semantically equivalent nodes are merged by clustering their embeddings with Affinity Propagation, which identifies an exemplar title for each group.

This process of abstraction and normalization ensures our leaf nodes are distinct and well-defined, forming a robust foundation for the hierarchy.

\subsection{Hierarchical Taxonomy Construction}\label{sec:hierarchical}  
With a solid set of leaf nodes, we build the taxonomy upwards. This is a complex reasoning task: grouping concepts from specific to general into a coherent hierarchy, similar to standard taxonomies. A single-pass LLM approach is prone to errors that propagate and corrupt the entire tree. 
We therefore employ a deliberate, level-by-level multi-agent framework with a ``Generator'' and an ``Evaluator'', to ensure coherence at each stage of construction. Starting with the leaf nodes (Level 0), the hierarchy is built using the following iterative process.

\noindent \emph{Generator}: An LLM proposes parent concepts to group the current level's nodes. This step performs the specific-to-general reasoning, outputting parent titles, descriptions, and child-parent mappings.

\noindent \emph{Evaluator}: To prevent error propagation, an Evaluator agent scrutinizes the Generator's output for logical consistency. It checks for common failure modes identified in our early explorations: missing child nodes, children assigned to multiple parents, and hallucinated mappings to non-existent nodes.

\noindent \emph{Generate-Evaluate Cycle}: If the Evaluator finds flaws, it provides feedback to the Generator to refine its output. This cycle repeats until the structure is validated. The new parent nodes then become the input for the next level, a process that continues until the number of newly generated parent nodes falls below a set threshold.

\section{Experiments}\label{sec:experiments}
To evaluate CLIMB's performance, we use three real-world datasets from different regions with varied sizes, comparing it against two baselines using a variety of evaluation metrics.

\textbf{Datasets.} We collected three datasets from major job search websites in their respective regions: \textbf{Palestine} with 2701 postings in English and Arabic, \textbf{Botswana} with 4854 English postings, and \textbf{USA} with 11285 English postings.
For each dataset, we randomly split the data into training and testing sets with a ratio of 9:1. The training set is used for generating the taxonomy, and the testing set is used for evaluating the quality of the taxonomy. More details are provided in Appendix~\ref{app:job-posting-datasets-construction}.

\textbf{Baselines.} We compare CLIMB against a state-of-the-art automated method and an expert-curated standard.
\textbf{TnT}~\cite{wan2024tnt}: As introduced in Section~\ref{sec:related-work}, TnT is our primary baseline as the state-of-the-art fully automated, bottom-up method. As its original prompts create a flat structure for user intents, we minimally adapted them for the occupation domain, as a full hierarchical adaptation is non-trivial and would require modifying its core logic.
\textbf{ESCO}~\cite{ESCO}: Built upon the international ISCO-08~\cite{ISCO08} standard, the European Skills, Competences, Qualifications and Occupations (ESCO) is a comprehensive, expert-curated taxonomy designed for \emph{region-agnostic} application. As the full taxonomy is too large to fit within the context window of the LLM annotators used for evaluation, we use a two-level version corresponding to its four- and five-digit codes, a necessary simplification.

\textbf{Evaluation Metrics.} To evaluate the taxonomies scalably, we employ a panel of three LLMs as independent annotators, a common and cost-effective strategy for reliable annotation \cite{gilardi2023chatgpt,zheng2023judging}. Our panel consists of Gemini 2.5 Flash, O4-Mini, and DeepSeek R1 (see Appendix \ref{app:implementation-details} for full model details). These annotators are tasked with labeling job postings from the test set using a given taxonomy. 
We then assess the quality of each taxonomy across three key dimensions: accuracy, comprehensiveness, and efficiency, using standard metrics from prior work \cite{shah2023using,wan2024tnt} applied to the LLM-generated labels:

\noindent \underline{\emph{Accuracy.}} High-quality taxonomies should be consistent (no contradictions), clear (unambiguous definitions), and accurate (correct categorizations). Without the ground truth labels for the job postings' occupations, we use the inter-annotator agreement as a proxy for label accuracy. 
For hierarchical taxonomies, we also measure the hierarchical agreement to provide a more detailed view. This gives us the following two metrics.
\textbf{Strict Agreement:} Requires all annotators to assign identical occupation labels to a job, and not choose ``Other.'' This strict metric evaluates label clarity and consistent interpretation.
\textbf{Hierarchical Agreement:} A more flexible metric for tree structures where agreement is met if assigned labels share an identical parent node at a given level. This rewards semantically close, if not exact, predictions.
We use \emph{Fleiss' kappa} \cite{fleiss1971measuring} to measure agreement, interpreting values using standard ranges \cite{LandisKoch1977, McHugh2012}: \(< 0.20\) (slight), \(0.21{-}0.40\) (fair), \(0.41{-}0.60\) (moderate), \(0.61{-}0.80\) (substantial), \(0.81{-}1.00\) (almost perfect).

\noindent \underline{\emph{Comprehensiveness.}} Measured by \textbf{Coverage Rate}, the percentage of jobs assigned a specific label rather than ``Other.'' A high rate indicates the taxonomy is sufficiently comprehensive for the corpus.

\noindent \underline{\emph{Efficiency.}} Measured by \textbf{Label Utilization Rate}, the percentage of labels used at least once during annotation, indicating how much of the taxonomy's breadth is relevant.
\begin{table}[h!]
    \centering
    \caption{Comparison of taxonomy sizes (number of nodes) against training dataset sizes. L0, L1, L2 refer to the levels of the taxonomy, from specific to general; `-' indicates a non-existent level. ESCO is a dataset-agnostic static taxonomy. 
    TnT generates lower number of nodes in general, and does not appear to scale with the size of the training dataset. CLIMB generates taxonomies that scale with the size of the training dataset.}
    \label{tab:taxonomy_size}
    \vspace{-0.2cm}
    \footnotesize
    \setlength{\tabcolsep}{4pt}
    \begin{tabular}{llcccc}
    \hline
    \textbf{Dataset (\# Jobs)} & \textbf{Taxonomy} & \textbf{L2} & \textbf{L1} & \textbf{L0} & \textbf{Total} \\
    \hline
    - & ESCO & - & 436 & 1760 & 2196 \\
    \hline
    \multirow{2}{*}{Palestine (2430)} & TnT & - & - & 87 & 87 \\
     & CLIMB & 14 & 32 & 91 & 143 \\
    \hline
    \multirow{2}{*}{Botswana (4368)} & TnT & - & - & 200 & 200 \\
     & CLIMB & 29 & 99 & 277 & 418 \\
    \hline
    \multirow{2}{*}{USA (10129)} & TnT & - & - & 130 & 130 \\
     & CLIMB & 190 & 341 & 671 & 1298 \\
    \hline
    \end{tabular}
\end{table}
\vspace{-0.2cm}
\begin{table}[h!]
    \centering
    \caption{Taxonomy comparison across datasets (agreement numbers are kappa values with interpretation symbols, others are percentages). Acronyms: Agr. (Agreement), Util. (average Label Utilization), Cov. (average Coverage Rate). Kappa interpretation: \textbf{*}= substantial, \textbf{**}= almost perfect. 
    $^\dagger$Agreement and Utilization at L2 for a fairer comparison between CLIMB and TnT where node numbers are similar (CLIMB: 190 vs. TnT: 130).}
    \label{tab:taxonomy_comparison}
    \vspace{-0.2cm}
    \footnotesize
    \setlength{\tabcolsep}{5pt}
    \begin{tabular}{llcccc}
    \hline
    \textbf{Dataset} & \textbf{Taxonomy} & \makecell{\textbf{Strict} \textbf{Agr.}} & \makecell{\textbf{Util.}} & \makecell{\textbf{Cov.}} \\
    \hline
    \multirow{3}{*}{Palestine} & ESCO & 0.46  & 5.24 & 95.33 \\
     & TnT & 0.59  & 48.66 & 89.30 \\
     & CLIMB & \textbf{0.73}\textbf{*}  & \textbf{54.50} & \textbf{98.52} \\
    \hline
    \multirow{3}{*}{Botswana} & ESCO & 0.56  & 10.02 & \textbf{98.83} \\
     & TnT & 0.42  & 38.00 & 81.21 \\
     & CLIMB & \textbf{0.63}\textbf{*}  & \textbf{44.86} & 97.81 \\
    \hline
    \multirow{4}{*}{USA} & ESCO & 0.46 & 17.28 & 97.20 \\
    & TnT & 0.62\textbf{*}  & \textbf{82.56} & 79.48 \\
    & CLIMB & 0.54  & 34.58 & \textbf{97.34} \\
    & CLIMB (L2)$^\dagger$ & \textbf{0.67}\textbf{*} & 82.28 & 97.34 \\
    \hline
    \end{tabular}
\end{table}
\begin{table}[h!]
    \centering
    \caption{Hierarchical agreement for CLIMB (kappa values), with levels numbered from most specific (L0) to most general. Kappa interpretation: \textbf{*}= substantial, \textbf{**}= almost perfect.}
    \label{tab:hierarchical_comparison}
    \vspace{-0.2cm}
    \footnotesize
    \setlength{\tabcolsep}{10pt}
    \begin{tabular}{llll}
    \hline
    \textbf{Dataset} & \textbf{Level 0} & \textbf{Level 1} & \textbf{Level 2} \\
    \hline
    Palestine & 0.73\textbf{*} & 0.81\textbf{**} & 0.82\textbf{**} \\
    Botswana & 0.63\textbf{*} & 0.72\textbf{*}& 0.76\textbf{*} \\
    USA & 0.54 & 0.66\textbf{*} & 0.67\textbf{*} \\
    \hline
    \end{tabular}
\end{table}

\section{Results}\label{sec:results}
We present quantitative (Section~\ref{sec:overall-performance-analysis}) and qualitative (Section~\ref{sec:qualitative-analysis}) results.

\subsection{Overall Performance \& Analysis}\label{sec:overall-performance-analysis}
We present our main quantitative results in this section. 
Table~\ref{tab:taxonomy_size} compares the sizes of the generated taxonomies against the data sizes used for generating them. For each dataset, we report the number of nodes at the levels that are used later for evaluation, where the total number of nodes is the sum of the nodes at all levels. The interactive visualizations of the trees generated by CLIMB are online \url{https://shorturl.at/LOkpT}. More descriptions are provided in Appendix~\ref{app:taxonomy-trees}.

Table~\ref{tab:taxonomy_comparison} evaluates the overall quality of the taxonomies using several metrics. To ensure a fair comparison on the USA dataset, this table includes an additional entry comparing taxonomies of similar size. Table~\ref{tab:hierarchical_comparison} assesses hierarchical agreement at different levels of the taxonomy generated by CLIMB, numbered bottom-up (Level 0 = leaves).

\begin{figure*}[h!]
    \centering
    \includegraphics[width=0.7\textwidth]{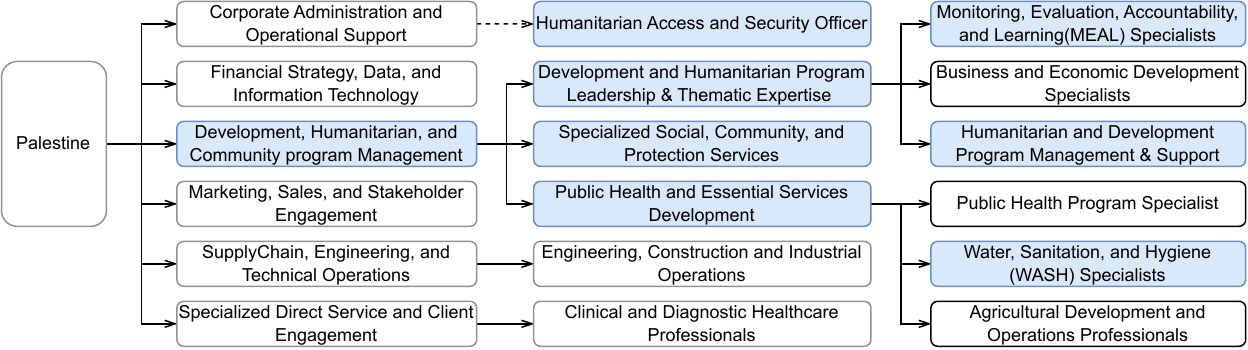}
    \vspace{0.2cm}

    \includegraphics[width=0.7\textwidth]{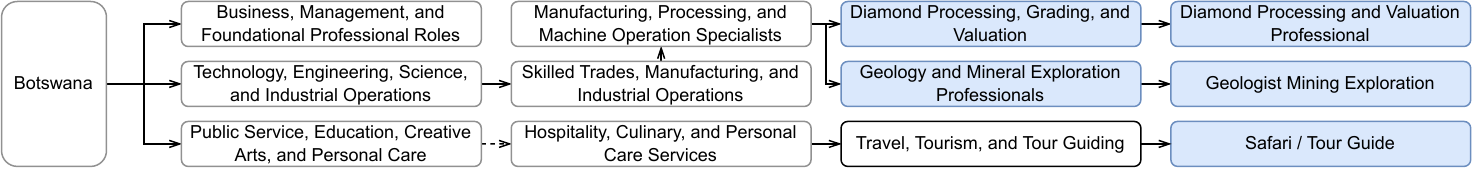}    
    \vspace{0.2cm}
    
    \includegraphics[width=0.7\textwidth]{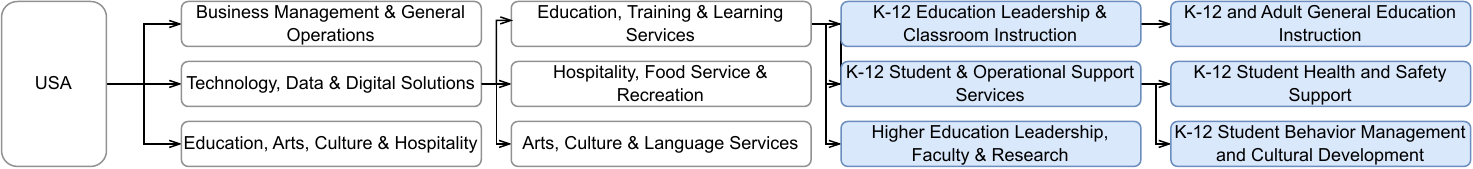}

    \caption{(a) A snippet from the Palestine taxonomy, showcasing how CLIMB identifies specialized humanitarian roles (e.g., MEAL Specialist) reflecting the local context. (b) A snippet from the Botswana taxonomy, showcasing how CLIMB reflects key economic drivers with roles in the diamond and tourism industries. (c) A snippet from the USA taxonomy, showing how CLIMB captures the specific structure of the American education system (K-12 vs. Higher Education). Dotted lines represent omitted intermediate nodes for clarity.}
    \label{fig:taxonomy_snippets}
\end{figure*}

Our main results demonstrate the effectiveness of CLIMB across multiple datasets.

\textbf{CLIMB generates significantly more consistent and unambiguous taxonomies}, leading to higher inter-annotator agreement. As shown in Table~\ref{tab:taxonomy_comparison}, CLIMB scores the highest in the Palestine and Botswana datasets with substantial agreement. While TnT appears to have a higher agreement score on the USA dataset, this direct comparison is misleading due to the vast taxonomy size difference (see Table~\ref{tab:taxonomy_size}). For a fairer comparison, we report CLIMB's agreement at Level 2, where the number of nodes (190) is much closer to TnT's (130). As shown in the last row for the USA dataset in Table~\ref{tab:taxonomy_comparison}, CLIMB's 0.67 agreement score surpasses TnT's 0.62, demonstrating superior structure, though its high granularity on this complex dataset reduces leaf-level clarity (Table~\ref{tab:hierarchical_comparison}).

\textbf{The hierarchy produced by CLIMB is logically structured and semantically coherent.} As shown in Table~\ref{tab:hierarchical_comparison}, agreement scores consistently increase as we ascend the hierarchy from more specific to broader categories, reaching almost perfect agreement for the Palestine dataset. This trend confirms that the generated structure is semantically coherent, as annotators tend to agree on broader concepts even when they diverge on finer-grained distinctions.

\textbf{CLIMB excels at producing a taxonomy that is both comprehensive and efficient}, striking a superior balance between coverage and utilization. In contrast, both baselines struggle with trading-off between these metrics. TnT achieves high label utilization only at the cost of comprehensiveness, failing to classify over 20\% of jobs in the USA dataset. Conversely, ESCO provides high coverage but suffers from extremely low utilization, indicating a bloated structure. 
CLIMB, however, delivers strong performance on both fronts. On the USA dataset, its leaf nodes provide excellent coverage (97.34\%). At a comparable size (L2), CLIMB achieves a similarly high utilization rate (82.28\% vs. 82.56\%) while achieving superior coverage.

\textbf{CLIMB generates taxonomies whose sizes scale logically with the size and complexity of the input corpus.} As shown in Table~\ref{tab:taxonomy_size},  the taxonomy generated by CLIMB for the large USA dataset is substantially larger than for the smaller Botswana and Palestine  datasets, reflecting the greater diversity of occupations. In contrast, the TnT baseline exhibits inconsistent scaling behavior, producing a smaller taxonomy for the largest dataset (130 nodes for USA) than for the mid-sized one (200 for Botswana). This suggests that CLIMB is more robust and sensitive to the underlying occupational diversity of the corpus, whereas TnT may fail to capture the full spectrum of occupations in larger, more complex datasets.

\subsection{Qualitative Analysis}\label{sec:qualitative-analysis}
Some unique structures can be found from the generated taxonomies for the USA, Palestine, and Botswana datasets, respectively. The qualitative analysis demonstrates CLIMB is indeed data-driven, and highlights CLIMB's ability to create taxonomies that reflect the unique socio-economic characteristics of each region, an advantage over any generic occupation taxonomy.

The taxonomy from the \textbf{Palestine dataset} (Figure~\ref{fig:taxonomy_snippets}(a)) reveals the prominence of humanitarian work, reflecting the region's socio-political context. The generated hierarchy dedicates one of its six top-level sectors to ``Development, Humanitarian, and Community Program Management,'' and related jobs appear across most other branches. This cross-cutting presence allows CLIMB to capture specialized roles like Monitoring, Evaluation, Accountability, and Learning (MEAL) Specialist and Water, Sanitation, and Hygiene (WASH) Specialist.

The taxonomy for the \textbf{Botswana dataset} (Figure~\ref{fig:taxonomy_snippets}(b)) mirrors the country's key economic drivers. It captures the granularity of its cornerstone diamond industry with specific roles like Diamond Processing, Grading, and Valuation. It also identifies occupations in the booming tourism sector, such as Safari Guide, which leverages the nation's natural beauty and wildlife.

For the \textbf{USA dataset} (Figure~\ref{fig:taxonomy_snippets}(c)), the CLIMB-generated taxonomy accurately models the American education system, distinctly structured into K-12 and higher education sectors. This level of regional specificity is absent in international standards like ESCO, demonstrating CLIMB's capacity to capture nuanced local labor market structures.

\section{Conclusion}\label{sec:conclusion}
In this work, we introduced CLIMB, a novel framework that automatically builds hierarchical occupation taxonomies directly from raw job postings. Our approach addresses the need for up-to-date taxonomies that fit specific regional job markets. It works from the bottom up, using semantic clustering to discover occupations and a multi-agent system to build a coherent hierarchy, all without needing seed terms or manual intervention.

Our experiments on three diverse, real-world datasets show that CLIMB outperforms existing methods, producing taxonomies that are demonstrably clearer, more coherent, and achieve a better balance of coverage and efficiency. Furthermore, our qualitative analysis confirms that the resulting taxonomies capture unique, regional labor market characteristics that generic models miss.

\newpage
\section*{Limitations}\label{sec:limitations}
Our study has several limitations that provide avenues for future work. \emph{First}, our methodology relied on LLMs as proxy HR experts for data annotation and as annotators for evaluation. While we used diverse, powerful models to mitigate risks, their judgments may contain biases or errors. A full validation against human experts for both the generated training data and the final taxonomy quality remains an area for future work. 

\emph{Second}, our use of Affinity Propagation for clustering, while effective on our datasets, has quadratic time complexity that may challenge scalability on much larger corpora; future work could explore alternatives like HDBSCAN. 

\emph{Third}, our comparison with the TnT baseline was based on a minimal adaptation of its original prompts. A more extensive modification of TnT to generate hierarchical structures, while desirable for a more direct comparison, was beyond the scope of this work due to its complexity and is left for future work. 

\emph{Finally}, the performance of CLIMB is inherently tied to the capabilities of the LLMs used. This includes the potential for semantic failures during hierarchy construction, where agents might create illogical groupings. Future advancements in LLM reasoning will likely improve taxonomy quality.

\section*{Acknowledgments}
The research leading to these results has received funding from the Special Research Fund (BOF) of Ghent University (BOF20/IBF/117), from the Flemish Government under the ``Onderzoeksprogramma Artificiële Intelligentie (AI) Vlaanderen'' programme, from the FWO (project no. G0F9816N, 3G042220, G073924N). Funded/Co-funded by the European Union (ERC, VIGILIA, 101142229). Views and opinions expressed are however those of the author(s) only and do not necessarily reflect those of the European Union or the European Research Council Executive Agency. Neither the European Union nor the granting authority can be held responsible for them. For the purpose of Open Access the author has applied a CC BY public copyright licence to any Author Accepted Manuscript version arising from this submission.

\bibliography{custom}

\appendix

\section{Job posting datasets construction}\label{app:job-posting-datasets-construction}

The three real-world datasets used in our experiments were constructed as follows:
\begin{itemize}[leftmargin=*]
    \item \textbf{Palestine Dataset}: Contains 2,701 job postings from \url{https://www.job.ps/}, a major job portal in Palestine. The data spans from September 2024 to February 2025.
    \item \textbf{Botswana Dataset}: Contains 4,854 job postings from \url{https://jobsbotswana.info/}, a leading job board in Botswana. The data covers the period from August 2023 to June 2025.
    \item \textbf{USA Dataset}: Contains 11,285 job postings from \url{https://www.indeed.com/}, with the location restricted to the United States. Data was collected between October 2023 and November 2023.
\end{itemize}

\section{Implementation Details}\label{app:implementation-details}
This section provides a detailed breakdown of each stage in the CLIMB pipeline, complementing the descriptions in Section~\ref{sec:method}.

\subsection{Job Posting Distillation \& Embedding}
The goal of this initial stage is to efficiently extract core occupational information from job postings, which are often verbose and contain irrelevant text (e.g., company boilerplate, application instructions). Applying a large language model (LLM) to summarize every posting in a large corpus would be prohibitively slow and expensive. To address this scalably, we adopt a two-step approach: first, we use a cost-effective LLM to generate a high-quality labeled dataset, and second, we train a lightweight yet effective classifier to perform the distillation on the entire corpus. This process ensures that the subsequent clustering stage operates on clean, relevant data.

\begin{itemize}[leftmargin=*]
    \item \textbf{Data Preparation}: Job descriptions first undergo basic cleaning to remove HTML tags and excess whitespace. They are then segmented into text chunks by paragraphs (split by new lines)

    \item \textbf{LLM-based Labeling for Training Data}: To create training data for the distillation classifier, we use \texttt{gpt-4o-mini} to annotate a sample of the text chunks. Each chunk is labeled as either relevant or irrelevant to defining the core occupation. The prompt guided the LLM to identify text segments directly describing job duties, responsibilities, and required qualifications, while ignoring other content. The full prompt details are available at \url{https://anonymous.4open.science/r/CLIMB/src/annotate_posting_segments.py}.

    \item \textbf{Distillation Classifier Training}: A lightweight binary classifier is trained on the LLM-annotated data to automate the relevance-filtering process for the entire corpus.
        \begin{itemize}[leftmargin=*]
            \item \textbf{Data Split}: The LLM-labeled dataset of text chunks was split into training (90\%) and testing (10\%) sets. The training set was further divided into training (90\%) and validation (10\%) subsets from training the classifier.
            \item \textbf{Model Architecture}: The classifier is an \texttt{XLMRobertaForSequenceClassification} model initialized with \texttt{BAAI/bge-m3} weights, featuring a classification head to output the binary prediction.
            \item \textbf{Training and Performance}: The model was trained for 1 epoch with a learning rate of 2e-5 and 100 warmup steps, using the AdamW optimizer. On the USA dataset, it achieved a test accuracy of 93.58\%, precision of 93.18\%, recall of 95.98\%, and an F1-score of 94.56\%, demonstrating its effectiveness in identifying relevant content.
        \end{itemize}

    \item \textbf{Distilled Description Generation}: After training, the classifier is applied to all text chunks in the corpus. For each job posting, the chunks classified as relevant are concatenated to form a ``distilled'' description.

    \item \textbf{Fallback Mechanism}: For robustness, if the distillation process results in no relevant text for a given posting (an infrequent event that occurred for only 9 jobs in the USA dataset), the full, preprocessed job description is used as a fallback to ensure no data is lost.

    \item \textbf{Embedding for Clustering}: Finally, the resulting distilled (or full) descriptions are embedded using the \texttt{Qwen3-Embedding-8B} model. This produces the final vector representations that serve as the input for the Semantic Clustering stage.
\end{itemize}

\subsection{Semantic Clustering}
This stage groups job postings into fine-grained clusters representing distinct occupations. The core idea is to train a custom similarity model that learns to mimic a human's nuanced judgment of what constitutes the ``same occupation,'' rather than relying on generic cosine similarity, which often falls short. This learned similarity is then used to drive a clustering algorithm.

\begin{itemize}[leftmargin=*]
    \item \textbf{Contrastive Data Sampling}: To train the similarity model, we first create a challenging set of job pairs. For each job in a dataset, we compute its cosine similarity with all other jobs using their embeddings. Based on this, we sample pairs to create a balanced mix of examples:
        \begin{itemize}[leftmargin=*]
            \item \textbf{Likely Easy Positives}: Two strongly similar jobs (from nearest neighbors ranked 1-20).
            \item \textbf{Likely Ambiguous Hard-Negatives or Hard-Positives}: One job with moderate similarity (from neighbors ranked 21-100).
            \item \textbf{Likely Easy Negatives}: One job with low similarity (from the rest of the corpus).
        \end{itemize}
        This strategy deliberately includes ambiguous (hard-negative or hard-positive) pairs to teach the model a more nuanced understanding. For the USA dataset, this process resulted in approximately 40,000 pairs for training.

    \item \textbf{LLM as HR Expert}: The sampled pairs are then annotated by \texttt{gpt-4o-mini}, which acts as a proxy for an HR expert. It labels each pair as ``same occupation'' or ``different occupation'' from the practical perspective of a job seeker. The full prompt is available at \url{https://anonymous.4open.science/r/CLIMB/src/same_occupation_job_pair_sampling.py}.

    \item \textbf{Similarity Classifier Training}: An XGBoost classifier is trained on the LLM-annotated pairs to predict the probability that two jobs belong to the same occupation.
        \begin{itemize}[leftmargin=*]
            \item \textbf{Features}: For a pair of job embeddings, \(e_a\) and \(e_b\), the input feature vector is the concatenation of \((e_a, e_b, e_a - e_b, e_a \odot e_b)\), where \(\odot\) denotes the Hadamard (element-wise) product. This feature engineering captures rich interaction information between the job descriptions.
            \item \textbf{Hyperparameters}: Key hyperparameters for the XGBoost model include a learning rate (\texttt{eta}) of 0.1, a \texttt{max\_depth} of 8, and a \texttt{subsample} ratio of 0.8. To account for class imbalance in the training data, \texttt{scale\_pos\_weight} was set to approximately 1.67. The model was trained with a maximum of 2000 boosting rounds, using early stopping with a patience of 50 rounds.
            \item \textbf{Training and Model Selection}: The dataset of labeled job pairs was split into training (90\%) and testing (10\%) sets. This setup was used to evaluate different embedding models for representing the jobs, including BGE-m3, Qwen3-Embedding-0.6B, Qwen3-Embedding-4B, and Qwen3-Embedding-8B. The Qwen3-Embedding-8B model was ultimately selected as it yielded the best classification performance on the test set.
        \end{itemize}

    \item \textbf{Similarity Matrix Construction}: After training, the XGBoost classifier is used to compute a pairwise similarity score for all jobs in the training set. To ensure the similarity matrix is symmetric (i.e., \(\text{sim}(A,B) = \text{sim}(B,A)\)), we define the final similarity between two jobs as the average of the two predictions: \((\text{score}(A,B) + \text{score}(B,A))/2\).
    
    \item \textbf{Clustering Algorithm Selection}: We explored several clustering algorithms, including Agglomerative Clustering, a custom greedy approach, and Affinity Propagation. Affinity Propagation was selected as it consistently yielded the best performance, as measured by the silhouette score on the generated clusters.

    \item \textbf{Final Clustering}: The Affinity Propagation algorithm is used to cluster the job postings based on the symmetrized similarity matrix. We used the default damping factor of 0.5 and set the preference hyperparameter to the median of the input similarities, which allows the algorithm to determine the number of clusters automatically.
\end{itemize}

\subsection{Leaf Node Generation}
This stage transforms the raw job posting clusters from the previous step into canonical, well-defined leaf nodes, each with a clear title and description, which serve as the foundation of the taxonomy. This involves two main sub-stages: abstracting a canonical occupation from each cluster's raw text and then refining the full set of generated occupations to ensure consistency and remove redundancy.

\begin{itemize}[leftmargin=*]
    \item \textbf{LLM-based Abstraction}: The goal of this step is to synthesize the content of each job cluster into a single, representative occupation.
    \begin{itemize}[leftmargin=*]
        \item \textbf{Model and Input}: For each cluster, we provide a sample of its raw job descriptions to the \texttt{gemini-2.5-flash-preview-05-20} model. Using a sample is necessary to manage the context window limits for clusters containing a large number of postings.
        \item \textbf{Prompting}: The LLM is prompted to act as an HR expert and generate a concise, generic title and a comprehensive description that canonically represents the occupation for the given job descriptions. The full prompt is available at \url{https://anonymous.4open.science/r/CLIMB/src/prompts.py}.
    \end{itemize}

    \item \textbf{Normalization and Deduplication}: The raw LLM-generated titles can be inconsistent or semantically redundant (e.g., ``Software Engineer'' vs. ``Software Developer''). To create a clean set of leaf nodes, we perform a three-step refinement process:
        \begin{itemize}[leftmargin=*]
            \item \textbf{Filtering}: Nodes with conjoined titles (e.g., ``Accountant + Bookkeeper'') are programmatically removed by detecting the ``+'' separator. These titles were intentionally generated by the LLM for clusters spanning multiple distinct occupations but are filtered out as they violate the principle of a node representing a single, distinct concept.
            \item \textbf{Cleaning}: Standard text normalization is applied to all titles for consistency. This includes converting text to lowercase, removing punctuation and stopwords, and performing lemmatization.
            \item \textbf{Deduplication}: To merge semantically equivalent nodes, we cluster them based on their meaning.
                \begin{enumerate}[leftmargin=*,label=\roman*.]
                    \item \textbf{Embedding}: The cleaned title and the full description of each node are embedded using the \texttt{Qwen3-Embedding-8B} model.
                    \item \textbf{Feature Representation}: A final vector for each node is created by concatenating its title embedding and description embedding. To emphasize the title's importance while retaining contextual information from the description, the title embedding is weighted at 80\% and the description embedding at 20\%.
                    \item \textbf{Clustering}: Affinity Propagation is applied to the cosine similarity matrix of these combined embeddings. The algorithm was configured with a damping factor of 0.7, and the preference was set to 0.95.
                    \item \textbf{Canonicalization}: For each resulting group of semantically similar nodes, the ``exemplar'' identified by Affinity Propagation is chosen as the single canonical representative, and the other nodes in the group are merged. This step ensures each distinct occupation is represented by only one node in the final set.
                \end{enumerate}
        \end{itemize}
\end{itemize}

\subsection{Hierarchical Taxonomy Construction}
This final stage constructs the taxonomy by building a hierarchy upwards from the canonical leaf nodes generated in the previous step. Constructing a deep and logically coherent hierarchy is a complex reasoning task. Our early explorations revealed that single-pass LLM approaches are prone to critical errors, such as inconsistent groupings or flawed parent-child relationships, which corrupt the entire structure. To address this, we designed a deliberate, level-by-level process using a reflection-based multi-agent framework. This ensures logical coherence at each step of the construction. The process is governed by two agents, a ``Generator'' and an ``Evaluator,'' operating in an iterative cycle.

\begin{itemize}[leftmargin=*]
    \item \textbf{Framework and Input}: The process begins with the set of canonical leaf nodes (Level 0), which are formatted as a JSON list of objects, each with a ``title'' and ``description''. The multi-agent framework iteratively processes this list to build the hierarchy one level at a time.

    \item \textbf{Generator Agent}:
        \begin{itemize}[leftmargin=*]
            \item \textbf{Model}: \texttt{gemini-2.5-pro-preview-05-06}. Specific generation hyperparameters (e.g., temperature) are detailed in the codebase, available at \url{https://anonymous.4open.science/r/CLIMB/src/tree_multiagent.py}.
            \item \textbf{Task}: At each level \(k\), the Generator's task is to take the set of nodes from that level and propose a set of parent concepts for the next level, \(k+1\). This involves performing specific-to-general abstraction to group the input concepts into broader parent categories.
            \item \textbf{Prompting and Output}: The agent uses a detailed prompt instructing it to return a JSON object containing the list of new parent nodes. For each parent, it must provide a concise ``title'', a comprehensive ``description'', and a list of the child nodes from level \(k\) that it subsumes. The prompt explicitly requires strict adherence to this JSON schema to ensure the output is machine-readable. It also includes a mechanism for receiving feedback from the Evaluator, allowing it to correct errors from previous attempts. The complete prompt templates are available at \url{https://anonymous.4open.science/r/CLIMB/src/tree_multiagent.py}.
        \end{itemize}

    \item \textbf{Evaluator Agent}: Unlike the Generator, the Evaluator is a deterministic, rule-based agent. It programmatically scrutinizes the Generator's output to ensure its logical coherence before it is accepted as a valid level in the hierarchy. It performs several critical checks:
        \begin{itemize}[leftmargin=*]
            \item \textbf{Completeness}: All child nodes from level \(k\) must be mapped to a parent in level \(k+1\).
            \item \textbf{Exclusivity}: No child node can be assigned to more than one parent, enforcing a strict tree structure.
            \item \textbf{Validity}: There are no ``hallucinated'' mappings to child nodes that did not exist in the input.
            \item \textbf{Constraints}: The number of generated parent nodes must be within a pre-defined range (see below).
        \end{itemize}

    \item \textbf{Iterative Generate-Evaluate Cycle}: The construction proceeds in a loop. If the Evaluator finds flaws in the Generator's output, its feedback is incorporated into the prompt for the next generation attempt. This cycle repeats until the output is fully validated. Once a level is validated, its newly generated parent nodes become the input for constructing the subsequent level. This process continues until the taxonomy converges to a small set of top-level categories, with the final hierarchy stored in a JSON file.

    \item \textbf{Grouping and Termination Constraints}: To guide the hierarchical construction, the process is bounded by the following constraints:
        \begin{itemize}[leftmargin=*]
            \item \textbf{Dynamic Grouping}: The number of parent nodes generated at each level is dynamic to adapt to the data's complexity. 
            \item \textbf{Termination}: The entire process terminates when either of two conditions is met: (1) the number of generated parent nodes at a new level is less than a threshold of 10, or (2) the number of nodes to be clustered in the current level is less than or equal to 1. This prevents the creation of overly granular or trivial top-level categories.
        \end{itemize}
\end{itemize}

\subsection{Evaluation}
To evaluate the quality of the generated taxonomies scalably and cost-effectively, we used a panel of three distinct LLMs as independent annotators. Each annotator was tasked with labeling every job posting in the test set according to a given taxonomy. This process is detailed below:
\begin{itemize}[leftmargin=*]
    \item \textbf{Annotator Panel}: The panel consisted of three models from different providers to ensure a diversity of perspectives: \texttt{o4-mini}, \texttt{gemini-2.5-flash-preview-04-17}, and \texttt{deepseek/deepseek-r1-0528}.
    \item \textbf{Annotation Task and Prompting}: For each job posting, an LLM was provided with the full job description and a string representation of one of the taxonomies (CLIMB, TnT, or ESCO). The prompt instructed the LLM to act as a job classification expert and adhere to several rules: (1) assign the most granular label possible, (2) use parent nodes for vague postings, (3) use multiple labels only for jobs that genuinely span distinct roles, and (4) use an ``Other'' category if no suitable label exists. The prompt also strictly enforced a JSON output format for programmatic parsing. The full prompt templates are available in the codebase at \url{https://anonymous.4open.science/r/CLIMB/src/taxonomy_evaluation.py}.
    \item \textbf{Execution and Parsing}: API calls were made with a temperature of 0.0 to ensure deterministic outputs. The final assigned labels for each job were then extracted from the returned JSON object to be used for calculating the evaluation metrics described in the main text.
\end{itemize}

\section{Taxonomy Trees}\label{app:taxonomy-trees}
The taxonomies generated by CLIMB for each dataset have the following hierarchical structures:
\begin{itemize}[leftmargin=*]
    \item \textbf{Palestine Dataset}: A 4-level taxonomy. The number of nodes per level, from the most general (top) to the most specific (bottom), is 6, 14, 32, and 91. The interactive visualization of the taxonomy is online \url{https://anonymous.4open.science/api/repo/CLIMB/file/demo/palestine.html}.
    \item \textbf{Botswana Dataset}: A 5-level taxonomy. The number of nodes per level, from most general to most specific, is 3, 10, 29, 99, and 277. The interactive visualization of the taxonomy is online \url{https://anonymous.4open.science/api/repo/CLIMB/file/demo/botswana.html}.
    \item \textbf{USA Dataset}: A 6-level taxonomy. The number of nodes per level, from most general to most specific, is 8, 23, 65, 190, 341, and 671. The interactive visualization of the taxonomy is online \url{https://anonymous.4open.science/api/repo/CLIMB/file/demo/usa.html}.
\end{itemize}

\end{document}